\documentclass[letterpaper, 10 pt, conference]{ieeeconf}  

\IEEEoverridecommandlockouts                              

\usepackage{graphicx} 
\usepackage{mathptmx} 
\usepackage{amsmath} 
\usepackage{amssymb}  
\usepackage{bm}
\usepackage{booktabs} 
\usepackage{subcaption}
\usepackage{cite}

\usepackage{lipsum}
\usepackage[normalem]{ulem}
\useunder{\uline}{\ul}{}

\usepackage{xcolor}
\newcommand{\newtext}{\color{black}}

\title{\LARGE \bf
M2R2: MultiModal Robotic Representation for Temporal Action Segmentation
}

\author{ Daniel Sliwowski$^{1}$, and Dongheui Lee$^{1,2}$
\thanks{This work was supported by the European Union project INVERSE under grant agreement No. 101136067 and in part supported by the Robot Industry Core Technology Development Program under Grant No. 00416440 funded by the Korea Ministry of Trade, Industry and Energy (MOTIE).}
\thanks{$^{1}$Daniel Sliwowski and Dongheui Lee are affiliated with the Autonomous Systems Lab, Technische Universität Wien (TU Wien), Vienna, Austria (e-mail: \texttt{\{daniel.sliwowski, dongheui.lee\}@tuwien.ac.at}).}%
\thanks{$^{2}$Dongheui Lee is also affiliated with the Institute of Robotics and Mechatronics, German Aerospace Center (DLR), Wessling, Germany.}%
}

\begin{document}

\maketitle
\thispagestyle{empty}
\pagestyle{empty}

\begin{abstract}
Temporal action segmentation (TAS) has long been a key area of research in both robotics and computer vision. In robotics, algorithms have primarily focused on leveraging proprioceptive information to determine skill boundaries, with recent approaches in surgical robotics incorporating vision. In contrast, computer vision typically relies on exteroceptive sensors, such as cameras. Existing multimodal TAS models in robotics integrate feature fusion within the model, making it difficult to reuse learned features across different models. Meanwhile, pretrained vision-only feature extractors commonly used in computer vision struggle in scenarios with limited object visibility. In this work, we address these challenges by proposing M2R2, a multimodal feature extractor tailored for TAS, which combines information from both proprioceptive and exteroceptive sensors. We introduce a novel training strategy that enables the reuse of learned features across multiple TAS models. Our method sets a new state-of-the-art performance on three robotic datasets REASSEMBLE, (Im)PerfectPour, and JIGSAWS. Additionally, we conduct an extensive ablation study to evaluate the contribution of different modalities in robotic TAS tasks.
\end{abstract}

\section{Introduction}
Temporal action segmentation (TAS) is a long-standing challenge in both robotics and computer vision. The goal is to identify temporal boundaries between actions in untrimmed data and assign action labels to each segment, depending on the subsequent tasks. In robotics, TAS helps break down complex demonstrations into simpler actions for learning motion policies~\cite{shi2023waypointbased}, or action preconditions and effects~\cite{sliwowski2024conditionnet}. In computer vision, TAS is applied to tasks like medical procedure documentation~\cite{TAS_survey}.

\begin{figure}
    \centering
    \includegraphics[width=\linewidth]{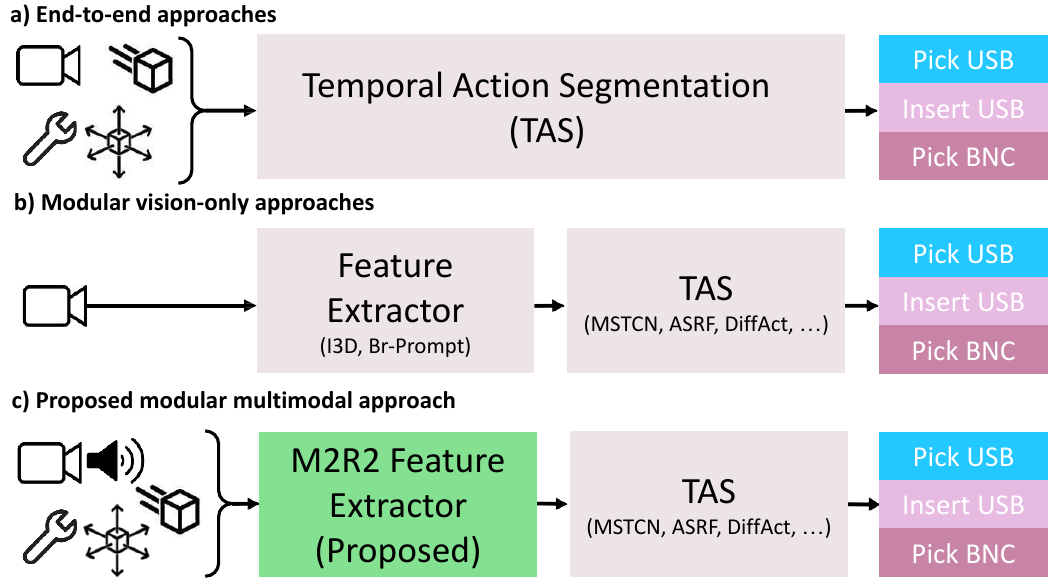}
    \caption{Many multimodal approaches often follow an end-to-end paradigm where the Temporal Action Segmentation (TAS) method operates directly on sensory inputs (a). In contrast, modular approaches train the feature extractor and TAS model separately, enabling component exchange but they usually rely on vision-only extractors (b). In this work, we propose the Multimodal Robotic Representation (M2R2) feature extractor, which leverages multimodal information and integrates with diverse TAS models (c).}
    \label{fig:Overview}
\end{figure}

In robotics, proprioceptive sensory information from the robot, such as forces, torques, joint positions, and end-effector poses and twists, are used to solve the TAS task. These data streams are analyzed to identify meaningful boundary points that indicate action transitions. Existing approaches often rely on heuristics, such as detecting change points in force and torque derivatives~\cite{BOCPD}, identifying sections of linear motion~\cite{shi2023waypointbased}, using handcrafted feature spaces combined with SVM classifiers~\cite{EIBAND2023104367}, or recognizing predefined skills in the demonstrations~\cite{DMP_seg}. In recent years, the field of surgical robotics has focused on leveraging vision and kinematic data for TAS, a task often referred to as surgical gesture recognition~\cite{TAS_surgery_surv}.  

On the other hand, the computer vision community primarily focuses on human temporal action segmentation, where proprioceptive data is unavailable. As a result, existing methods rely on exteroceptive sensors such as videos~\cite{mstcn}, optical flow~\cite{mstcn}. Pretrained feature extractors are first used to obtain meaningful latent representations of the observations, which serve as inputs to multi-stage prediction models. These models iteratively refine action and boundary predictions at each stage~\cite{mstcn}. Prior research has shown that deep learning-based TAS models can effectively learn meaningful feature representations and decompose complex, long-horizon demonstrations into simpler action sequences~\cite{mstcn, asrf, diffact}.  

Despite promising advances in leveraging vision and multimodal data for robotic TAS in surgical robotics, proprioceptive-based approaches remain the predominant choice for larger scale manipulation tasks such as assembly and disassembly. However, these methods often rely on handcrafted heuristics, requiring extensive hyperparameter tuning or the construction of a predefined skill library before segmentation~\cite{DMP_seg, BOCPD}. In contrast, purely vision-based approaches struggle in scenarios with occlusions or visually similar objects. For example, distinguishing between two pegs that differ by only a few millimeters is particularly challenging due to their small size, yet such precision is crucial for learning accurate insertion policies. While multimodal approaches have shown promise in surgical robotics, they are typically trained end-to-end, restricting flexibility in modifying the action segmentation head. This limitation makes it difficult to incorporate state-of-the-art TAS models or adapt to new models as they emerge. In contrast, separating feature fusion and extraction from the TAS model allows for seamless integration of improved TAS architectures in the future.

To address these challenges, in this work, we develop a multimodal deep learning feature extractor and a {\newtext training} strategy which is tailored for robotic temporal action segmentation tasks, and can be matched with any TAS model, which fosters easier reusability of the learned features. We refer to the resulting fused features as multimodal robotic representation (M2R2) features. To compute the M2R2 features, we leverage proprioceptive information such as force and torque measurements, end-effector pose and twist, and gripper width. For exteroceptive data, we use RGB cameras and audio. In this work, we adopt a late fusion strategy, where each modality is first processed independently and then fused using a transformer-based model~\cite{vaswani2017attention}. In our experiments, the resulting M2R2 features improve temporal action segmentation performance over existing end-to-end and vision-only approaches. {\newtext Moreover, M2R2 features can be used successfully with various state-of-the-art TAS models and that the framework generalizes well across different embodiments and task domains.} A general overview of the proposed approach is shown in Figure~\ref{fig:Overview}.  

To summarize, our main contributions are as follows:
\begin{enumerate}
    \item A deep-learning-based multimodal feature extractor for robotic temporal action segmentation.
    \item A {\newtext training} strategy for learning multimodal features for robotic temporal action segmentation.
    \item An extensive evaluation of the influence of different sensor modalities on robotic temporal action segmentation performance.
\end{enumerate}

\section{Related works}

\subsection{Deep Learning TAS}
Many deep learning TAS models use pretrained visual features instead of training end-to-end~\cite{mstcn, asrf, diffact}. These feature extractors must capture both short- and long-term dependencies in the data. Over the years, several video feature extractors have been proposed. Carreira et al. introduced I3D features~\cite{I3D_features}, leveraging both RGB and optical flow in a two-stream 3D convolution network. More recently, Wang et al. introduced ActionCLIP~\cite{ActionCLIP}, which uses a contrastive text-to-video pretraining strategy to extend CLIP image features to video classification. However, these features are not well-suited for TAS, as they were trained on trimmed action clips. To address this, Li et al. developed Br-Prompt~\cite{BRPrompt}, by extending ActionCLIP to understand action execution order. 

In the second stage, TAS models are trained on the extracted features. Farha et al. introduced MSTCN~\cite{mstcn} to reduce over-segmentation by refining predictions across multiple stages. Alternatively, Ishikawa et al. proposed the ASRF model~\cite{asrf}, which regresses temporal action boundaries in addition to per-frame class predictions. More recently, models have incorporated modern architectures, including diffusion models~\cite{diffact}. Additionally, some approaches refine temporal segmentation by performing hierarchical refinement on the predicted action segments~\cite{ahn2021refining}.

Although much of the deep learning TAS literature focuses on human action segmentation, where proprioceptive data is typically unavailable, robotic TAS can leverage proprioceptive signals to improve segmentation accuracy. In this work, we extend techniques for learning robust video feature representations to fuse both proprioceptive and exteroceptive data, resulting in improved segmentation performance. We further show that these multimodal features can be seamlessly integrated into existing robotic TAS models without requiring changes to their architecture.


\subsection{Robotics TAS}
In recent years, surgical robotics has increasingly focused on vision-based and multimodal TAS, also known as surgical gesture recognition. Earlier methods used Convolutional Neural Networks (CNNs) for action segmentation from vision data~\cite{lea2016segmental}. More recent approaches employ transformer models for merging proprioceptive and vision information~\cite{weerasinghe2024multimodal}. However, existing multimodal TAS models merge feature extraction, and temporal action segmentation into a single end-to-end model, limiting the reusability of learned feature extractors across different TAS architectures. In contrast, decoupling feature extraction from the TAS model allows the learned features to be reused with various TAS models.

Despite the success of multimodal TAS in surgical robotics, proprioception-only TAS remains the primary method for robotic manipulation tasks, such as assembly and disassembly. Proprioception-based TAS can be classified into supervised and unsupervised methods. Supervised methods rely on labeled data to model primitive actions for recognizing actions in new task demonstrations. Early approaches~\cite{DMP_seg} assume access to a skill library. More recent methods segment skills by comparing the current scene state to a symbolic representation of known skills and using a Support Vector Machine (SVM) to classify skills within a sliding window. These methods rely on handcrafted features such as power and contact duration~\cite{EIBAND2023104367}.

\begin{figure*}
    \centering
    \includegraphics[width=\linewidth]{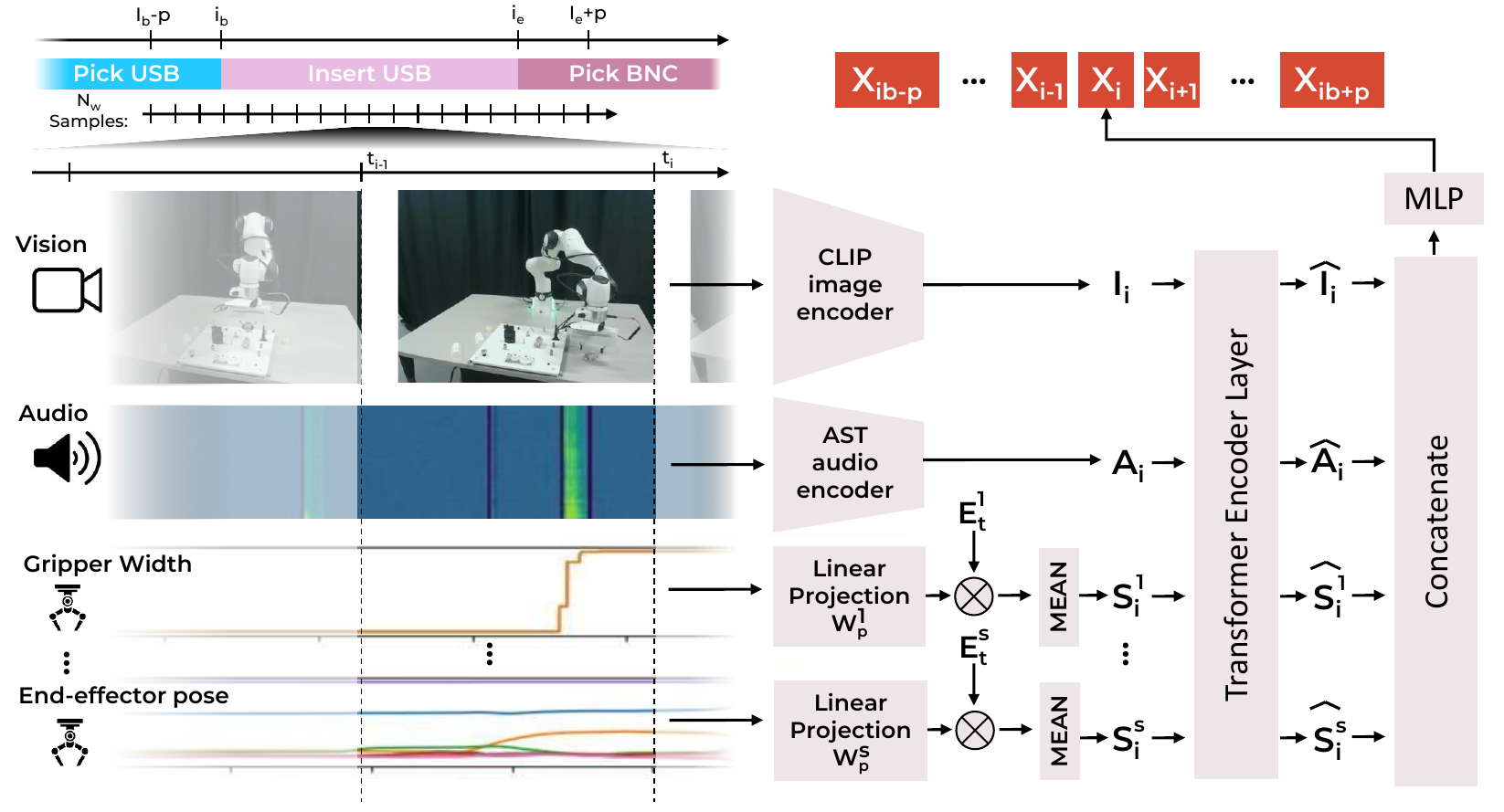}
    \caption{\textbf{M2R2 Feature Extractor Architecture.} To compute the multimodal feature at time instant $t_i$, we first process each modality separately to obtain image features $I_i$, audio features $A_i$, and proprioceptive features $\{S_i^s\}_{s=1}^{N_s}$, which are later fused using a Transformer encoder layer followed by an MLP. To obtain $I_i$, we use the ActionCLIP image encoder~\cite{ActionCLIP}. For $A_i$, we extract features using the Audio Spectrogram Transformer~\cite{gong2021ast}. For the proprioceptive data, we compute $S_i^s$ by first upscaling the raw sensory data through a linear projection into a higher-dimensional space using a learnable projection matrix $W^s_p$. Next, we embed temporal information by applying an element-wise multiplication with a learnable temporal embedding matrix $E^s_t$. Finally, we compute the average over the temporal dimension to obtain $S_i^s$.}
    \label{fig:archi}
\end{figure*}

Unsupervised methods detect action boundaries without predefined skill models. Probabilistic approaches apply Gaussian Mixture Models (GMMs)~\cite{krishnan2017transition} or HMMs~\cite{grigore2017discovering} to infer segmentation points. More recent work uses Bayesian Online Changepoint Detection (BOCPD) to find transitions in noisy force and torque data~\cite{BOCPD}. Other methods model segments as linear trajectories, detecting boundaries from reconstruction error against a threshold~\cite{shi2023waypointbased}.

Despite these advances, current TAS models in robotics face limitations. Many rely on predefined skill libraries, which require prior knowledge and limit generalization~\cite{DMP_seg, EIBAND2023104367}. Others depend on handcrafted features that do not scale well to diverse tasks~\cite{weerasinghe2024multimodal}. Many also need heavy hyperparameter tuning, making adaptation to new environments difficult~\cite{BOCPD, shi2023waypointbased}. In contrast, M2R2 processes raw sensory data with a deep neural network that learns task-relevant features. By decoupling feature extraction from the TAS model, M2R2 improves reusability and integrates seamlessly with different models.



\section{Multimodal Feature Extraction and Training}
\subsection{Problem formulation}
\begin{figure}
    \centering
    \includegraphics[width=\linewidth]{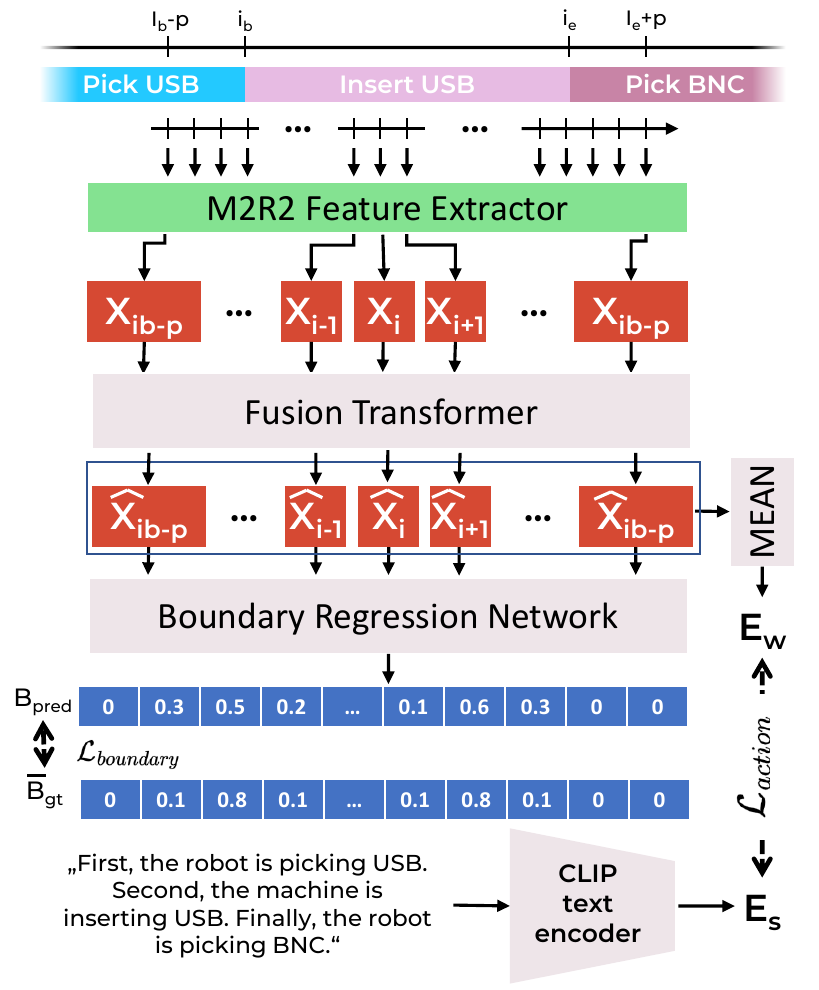}
    \caption{\textbf{M2R2 training objectives.} Given a window $[i_b - p, i_e + p)$, we sample $N_w$ frames and extract features using our M2R2 feature extractor. A Temporal Fusion Transformer refines these features into $\widehat{X}$, which we average to obtain the window representation $E_w$. To learn action order, we minimize the distance between $E_w$ and a textual embedding $E_s$ generated from action labels by using a template. To enhance boundary detection, we minimize the MSE between the smoothed ground truth boundaries $\bar{B}_{gt}$ (obtained from the frame-wise dataset annotations) and predictions $B_{pred}$, obtained via a Boundary Regression Network.}
    \label{fig:pretrain}
\end{figure}

The goal of our work is to learn a multimodal feature extractor $F_m$ for temporal action segmentation that fuses exteroceptive and proprioceptive data. Given visual observations $V = \{v_i\}_{i=1}^{T_{image}}$, audio $A = \{a_i\}_{i=1}^{T_{audio}}$, and proprioception $P$, we extract multimodal features $X = \{x_i\}_{i=1...T}$ such that $X = F_m(V, A, P)$, where
\begin{equation}
    P = \bigcup_{k=1}^{N_s} \left\{ \left( t_{k,i}, s_k(t_{k,i}) \right) \mid t_{k,i} \in T_k \right\}
\end{equation}
represents the collection of $N_s$ proprioceptive sensor modalities, where $t_{k,i}$ is the timestamp of the $i$-th measurement from sensor $k$, and $T_k$ is the set of timestamps for sensor $k$. As sensors operate at different frequencies, features are fused to align with the slowest sensor, here the camera. The extracted features $X$ are then used to predict action segmentation labels $C = \{c_i\}_{i=1}^{T}$ with a state-of-the-art TAS model $F_{seg}$, such that $C = F_{seg}(X)$. Figure~\ref{fig:archi} shows fusion of different modalities at a single time step $i$ to obtain $x_i$, while Figure~\ref{fig:pretrain} shows how these features are processed for {\newtext training} the M2R2 extractor.

\subsection{Preprocessing}
Different sensors operate at varying frequencies: standard RGB cameras run at 10–60 Hz, robot proprioceptive sensors at 100 Hz–1 kHz, and microphones above 16 kHz. Simply downsampling high-frequency data to match the slowest sensor can result in the loss of important information. For instance, in contact-rich manipulation, rapid force changes during contact or high-pitched audio clicks may be missed.  To address this, we preprocess data as follows. For a camera frame at $t_i$ and the previous frame at $t_{i-1}$, we collect proprioceptive measurements in $[t_{i-1}, t_i)$ and resample them to a common frequency $f$ below all sensor rates using linear interpolation; we choose $f = 300$ Hz. Finally, we compute the mean and standard deviation of each sensor over the training set for normalization.

For audio signals, we first resample the sensor readings within the window to 16 kHz. Following previous works, we compute the audio spectrogram using the mel filter banks~\cite{gong2021ast}. We set the window size to 400 samples and use 64 mel filter banks. Finally, we compute the logarithm of the resulting spectrogram and normalize it to the range $[-1,1]$ using the minimum and maximum log-spectrogram values over the entire training dataset.

\subsection{M2R2 Feature Extraction}
Instead of using the full demonstration, we sample shorter data windows, each containing three consecutive actions to preserve local relationships. For each action segment with start and end indices $[i_b, i_e)$, we expand the boundaries by a padding $p = 30$ to include contextual information from neighboring actions, resulting in $[i_b - p, i_e + p)$. We then randomly sample $N_w = 100$ points within these extended boundaries. To ensure adequate representation of each action the number of samples in each section, $[i_b - p, i_b)$, $[i_b, i_e)$, $[i_e, i_e + p)$, is proportional to its duration. Each window also includes the natural language descriptions of the contained actions and a boundary sequence $B_{gt} \in \{0,1\}^{N_w}$ marking action transition points within the window. To account for smooth transitions, we apply a Gaussian kernel to obtain a soft boundary representation, $\bar{B}_{gt} = \lambda(B_{gt})$. We use a kernel width of 1.

Next, our goal is to obtain the multimodal M2R2 feature $x_i$ for each sample within the window. To this end, we first compute a single representation for each sensor modality corresponding to sample $x_i$, which we then fuse using a self-attention modality transformer model~\cite{vaswani2017attention}. For simplicity, we describe the procedure for a single sample, but this process is repeated for all sampled data points.

\noindent
\textbf{Images:} We use a Visual Transformer initialized with ActionCLIP weights~\cite{ActionCLIP}, to produce the image representation $I_i \in \mathbb{R}^{D_e}$, where $D_e = 512$ is the M2R2 embedding size.

\noindent
\textbf{Audio:} We use the Audio Spectrogram Transformer (AST)~\cite{gong2021ast} to extract audio features from the log-mel spectrogram. The classification token from the AST's last hidden layer gives the audio representation $A_i \in \mathbb{R}^{D_e}$.

\noindent
\textbf{Proprioceptive Sensors:} We use a learnable temporal fusion approach to compute each sensor's representation. After preprocessing, sensor $s$ produces a time series of shape $D_s \times T_s$, where $D_s$ is the feature dimension of the sensor and $T_s$ the number of temporal samples. At time step $i$, we project the sensor data into a $D_e$-dimensional space using a learned matrix $W_p^s$, we then apply temporal fusion via a learnable embedding $E_t^s \in \mathbb{R}^{D_e \times T_s}$ with element-wise multiplication. Averaging over time yields the sensor representation $S_i^s \in \mathbb{R}^{D_e}$. Each sensor has its own projection and embedding matrices.

Finally, we fuse information from all sensor modalities using a modality transformer followed by a multi-layer perceptron (MLP), similar to~\cite{li2022see}. We first construct a sequence of tokens by stacking the individual modality representations:
\begin{equation}
    \text{tok} = [I_i, A_i, S^1_i, ..., S^s_i] \in \mathbb{R}^{D_e \times (N_s+2)}.
\end{equation}
Next, we exchange information between modalities using a single self-attention transformer encoder layer~\cite{vaswani2017attention}. This operation results in updated modality representations:
\begin{equation}
    \widehat{\text{tok}} = [\widehat{I_i}, \widehat{A_i}, \widehat{S^1_i}, ..., \widehat{S^s_i}].
\end{equation}
We then fuse all tokens by concatenating them into a single vector of shape $(1 \times D_e \cdot (N_s+2))$, which is passed through an MLP to obtain the final multimodal representation: $x_i \in \mathbb{R}^{D_e}.$

\subsection{M2R2 {\newtext training}}
We adopt a {\newtext training} strategy similar to Br-Prompt~\cite{li2022see}. To capture temporal dependencies between the multimodal features computed for each sampled datapoint in the window, we use a multi-head self-attention transformer encoder with $L$ layers, which we refer to as the \textit{Fusion Transformer}. The Fusion Transformer takes as input the sequence of multimodal features $X \in \mathbb{R}^{T \times D_e}$ and produces a temporally enhanced sequence of features $\widehat{X} \in \mathbb{R}^{T \times D_e}$. To obtain a representation of the entire window, we compute the mean value over the temporal dimension, given by $E_w = \frac{1}{T} \sum_{i=1}^{T} \widehat{X}_i$, where $E_w \in \mathbb{R}^{D_e}$. Additionally, we input $\widehat{X}$ into an MLP, referred to as the \textit{Boundary Regression Network}. The purpose of this network is to predict the probability of each sample within the window being a boundary sample. This encourages the model to learn more meaningful action boundaries from the underlying multimodal data. We denote the predicted boundary sequence as $B_{\text{pred}} \in \mathbb{R}^{T}$.

To {\newtext train} the M2R2 feature extractor, we use two training objectives. First, we maximize the similarity of the window representation $E_w$ and an embedding $E_s$ of a sentence describing the order of actions (e.g. ``\textit{First, the robot does action A. Next, it performs action B. Finally, the machine executes action C}"). We obtain this sentence embedding by first tokenizing the sentence and then using a transformer model initialized with ActionCLIP weights. Next, we maximize the similarity between the window embedding $E_w$ and its corresponding sentence embedding $E_s$, while minimizing the similarity to sentence embeddings of actions performed in different orders. To achieve this, we use the same loss function as presented in~\cite{BRPrompt}, which we denote as $\mathcal{L}_{\text{action}}$. 

Additionally, to learn the correct boundaries, we minimize the average Mean Squared Error between the predicted and ground truth boundary sequences over the entire batch:
\begin{equation}
    \mathcal{L}_{\text{boundary}} = \frac{1}{B_s} \sum_{i=1}^{B_s} \frac{1}{T} \sum_{t=1}^{T} (\bar{B}_{gt}^i - B_{\text{pred}}^i)^2,
\end{equation}
where $B_s$ is the batch size. Our final loss is the sum of both loss functions:
\begin{equation}
    \mathcal{L}_{\text{total}} = \mathcal{L}_{\text{action}} + \mathcal{L}_{\text{boundary}}.
\end{equation}

{
\newtext

\subsection{TAS model training}
After training the M2R2 feature extractor, we discard the fusion transformer, boundary regression network, and CLIP text encoder, as they are not needed for computing the per-frame representations. We then extract M2R2 features for each recording by first preprocessing the raw data as during training, and then passing each frame through the M2R2 extractor. For a recording, this yields a feature set $X \in \mathbb{R}^{T \times D_e}$, where $T$ is the number of frames and $D_e$ is the M2R2 feature dimension. This produces a preprocessed dataset $\mathcal{D} = \{X_i, C_i\}_i^{N}$, where $C_i$ are the per-frame ground-truth action labels for recording $i$. Finally, we use this dataset to train a temporal action segmentation model; for example, MSTCN~\cite{mstcn}, ASRF~\cite{asrf}, and DiffAct~\cite{diffact}.
}

\section{Experiments}
We conduct several experiments to evaluate the performance of the proposed M2R2 multimodal feature extractor. First, we compare the performance of the multimodal M2R2 features to several state-of-the-art temporal action segmentation approaches. {\newtext Next, we evaluate the generalization of the proposed M2R2 feature extractor to new datasets. Finally,} we conduct extensive ablation studies to assess the impact of each modality on the robustness of the learned features.

\subsection{Datasets}
{\newtext

For our experiments, we use the REASSEMBLE dataset~\cite{sliwowski2025reassemble}, which contains robot demonstrations of assembly and disassembly of connectors, gears, pegs, and nuts. The dataset consists of 149 recordings with a median of 36 actions per recording and strong visual similarity between objects, making temporal action segmentation particularly challenging. We use the external cameras, the gripper-mounted microphone, force/torque sensor, and end-effector pose and twist to compute the M2R2 features.  In total, the REASSEMBLE dataset contains 69 fine-grain labels (activity+object, e.g. \textit{Insert USB}) and 4 coarse-level labels where the object names are discarded (activity, e.g. \textit{Insert}). 

We also use two additional datasets: (Im)PerfectPour~\cite{sliwowski2024conditionnet} and JIGSAWS~\cite{gao2014jhu}. (Im)PerfectPour contains 544 demonstrations of a robot performing bartending tasks. Since the available modalities (pose, twist, force, torque, and gripper width) are similar to REASSEMBLE, this dataset allows us to evaluate how features trained on one dataset generalize to a new task domain. In place of audio, all input values are set to $-1$ to indicate the absence of audio signals.

JIGSAWS is a surgical gesture recognition dataset. Following prior work, we use the Suturing task with a Leave-One-User-Out evaluation. The available modalities (pose, twist, and gripper width of both arms) differ significantly from REASSEMBLE; therefore, we train M2R2 using the available modalities. This dataset allows us to evaluate our approach on a new embodiment and domain.
}

\subsection{Metrics}
{\newtext
We evaluate temporal action segmentation (TAS) models using standard metrics~\cite{diffact}: frame-level accuracy, EDIT score, and segment F1 at 10\%, 25\%, and 50\% overlap. Frame-level accuracy measures correctly classified frames, EDIT score captures sequence alignment, and F1 scores assess segmentation quality under different overlap thresholds. 

In robotics, TAS research also reports detection rate (DR)~\cite{BOCPD, EIBAND2023104367}. DR is computed as an F1 score with a $\pm t_e$-frame window around each ground-truth boundary. A prediction inside the window counts as one true positive; others are false positives, and missed boundaries are false negatives.  We set $t_e = 10$ for REASSEMBLE and (Im)PerfectPour, and $t_e = 30$ for JIGSAWS, which corresponds to 1\,s of wall time.
}

\subsection{Baselines}
{\newtext For video feature extraction, two commonly used models in TAS are I3D~\cite{I3D_features} and Br-Prompt~\cite{BRPrompt}. We adopt Br-Prompt as our baseline vision-only extractor, as it achieves state-of-the-art performance on vision-based TAS tasks~\cite{BRPrompt}.} We {\newtext train} Br-Prompt with its default hyperparameters, sampling 16-frame windows during training. These windows typically cover 2--3 actions and, at most, 6. For comparison, we include two state-of-the-art proprioception-based approaches: BOCPD~\cite{BOCPD}, which segments actions from the 6DoF wrench, and AWE~\cite{shi2023waypointbased}, which relies on position trajectories.  We tune their hyperparameters to ensure the number of predicted segments aligns with the ground truth. Since these methods do not predict action labels, each segment is assigned the ground-truth class with the highest overlap. In addition, we consider three deep learning TAS models, MSTCN~\cite{mstcn}, ASRF~\cite{asrf}, and DiffAct~\cite{diffact}.

\begin{table*}[!t]
\caption{Quantitative results on the REASSEMBLE dataset~\cite{sliwowski2025reassemble} for different segmentation models: MSTCN~\cite{mstcn}, ASRF~\cite{asrf}, and DiffAct~\cite{diffact}. The best results are shown in \textbf{bold}, and the second-best are \underline{underlined}. BRP refers to Br-Prompt~\cite{BRPrompt}. V -- vision, A -- audio, P--pose, FT -- force and torque, G -- gripper width,  E2E -- end-to-end.}
\label{tab:quantitative}
\centering
\setlength{\tabcolsep}{3pt}
\begin{tabular}{llllcccccccccccc}
\toprule
&&&& \multicolumn{6}{c}{Fine-grain Labels} & \multicolumn{6}{c}{Coarse Label} \\
\cmidrule(lr){5-10} \cmidrule(lr){11-16}
Method & Prediction & Modalities & Type & \multicolumn{3}{c}{F1@\{10,25,50\}} & Acc & EDIT & DR & \multicolumn{3}{c}{F1@\{10,25,50\}} & Acc & EDIT & DR \\
\midrule
BOCPD~\cite{BOCPD}              & Boundary & FT & E2E & 39.2 & 32.6 & 12.8 & 35.0 & 19.7 & 17.4 & -- & -- & -- & -- & -- & -- \\
AWE~\cite{shi2023waypointbased} & Boundary & P & E2E & 72.5 & 68.4 & 35.8 & 54.6 & 54.9 & 22.1 & -- & -- & -- & -- & -- & -- \\
BRP+MSTCN                       & Boundary and Action & V & Modular & 12.9 & 10.9 & 8.4 & 7.9 & 18.4 & 18.5 & 32.1 & 28.2 & 20.0 & 65.0 & 34.0 & 12.2 \\
BRP+ASRF                        & Boundary and Action & V & Modular & 12.8 & 11.1 & 8.6 & 6.4 & 19.4 & 24.8 & 57.6 & 44.5 & 23.3 & 54.3 & 67.5 & 22.2                     \\
BRP+DiffAct                     & Boundary and Action & V & Modular & 12.0 & 10.1 & 5.1 & 5.7 & 21.4 & 19.9 & 62.9 & 51.8 & 22.5 & 54.8 & 73.4 & 17.9 \\
M2R2+MSTCN                      & Boundary and Action & V,A,P,FT,G & Modular & \underline{83.1} & \underline{82.7} & \underline{80.8} & \underline{82.4} & \underline{79.3} & \underline{89.5} & \underline{97.7} & \underline{97.6} & \underline{96.7} & \textbf{96.7} & \underline{95.0} & \underline{94.5} \\
M2R2+ASRF                       & Boundary and Action & V,A,P,FT,G & Modular & \textbf{83.5} & \textbf{83.4} & \textbf{82.4} & \textbf{82.7} & \textbf{82.5} & \textbf{95.1} & \textbf{98.1} & \textbf{98.1} & \textbf{97.1} & \underline{96.0} & \textbf{96.9} & \textbf{95.9} \\
M2R2+DiffAct                    & Boundary and Action & V,A,P,FT,G & Modular & 78.1 & 77.7 & 74.6 & 74.9 & 68.7 & 81.4 & 95.1 & 95.0 & 92.1 & 91.7 & 90.3 & 83.9 \\
\bottomrule
\end{tabular}
\end{table*}

\subsection{Quantitative Evaluation}
Table~\ref{tab:quantitative} presents the performance of each baseline model. We observe that deep learning-based TAS models trained with the multimodal M2R2 features achieve the best performance among all models. Specifically, M2R2 with ASRF achieves the highest performance with our features, attaining an F1@50 score of 82.4\%, surpassing the unsupervised robotic TAS models, by at least 46.6 percentage points.

\begin{figure*}
    \centering
    \includegraphics[width=\linewidth]{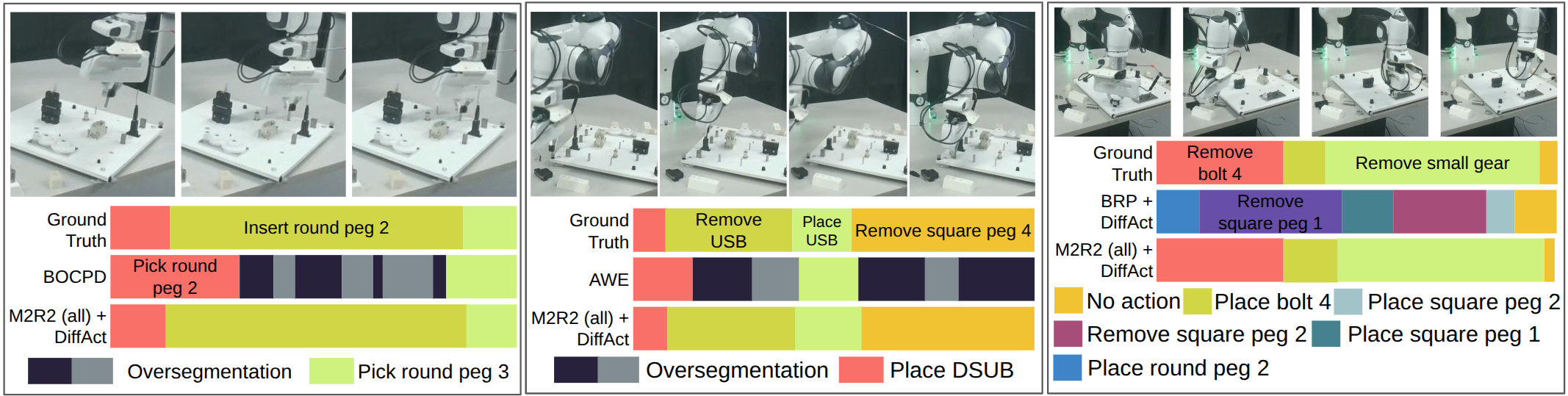}
    \caption{\newtext Qualitative evaluation of different baseline TAS models. The proposed M2R2 feature extractor is less sensitive to changes in proprioceptive data compared to heuristic approaches (left, middle) and more precise in differentiating objects than vision-only approaches (right).}
    \label{fig:qual}
\end{figure*}

Among all cases, the TAS trained with vision-only features perform the worst, with F1@50 scores {\newtext ranging from 5.1\% to 8.4\% depending on the TAS model. TAS models trained with vision-only features frequently confuse object categories, as shown in Figure~\ref{fig:qual}, because many REASSEMBLE objects are poorly visible due to their small size. Table~\ref{tab:ipp_quant} shows that vision-only features perform better when objects are more visible, as in the (Im)PerfectPour dataset. For both datasets, performance improves when adding proprioceptive information, underscoring its importance for robotic TAS.}

\begin{table}[t]
\caption{Quantitative results on the (Im)PerfectPour dataset~\cite{sliwowski2024conditionnet}. Best performing models in \textbf{bold}, second best are ~\underline{underlined}. M2R2 features trained on REASSEMBLE dataset~\cite{sliwowski2025reassemble} with all modalities. V -- vision, P--pose, FT -- force and torque, G -- gripper width.}
\label{tab:ipp_quant}
\centering
\setlength{\tabcolsep}{4pt}
\begin{tabular}{llllllll}
\toprule
Method    & Modalities  & \multicolumn{3}{c}{F1@\{10,25,50\}}                    & Acc              & EDIT   & DR          \\
\midrule
BRP+MSTCN      &  V    & 81.0          & 73.9          & 63.6          & 76.8          & 88.9                & 28.8          \\
BRP+ASRF       &  V    & 84.1          & 82.1          & 73.6          & 77.7          & \textbf{91.2}       & 28.3          \\
BRP+DiffAct    &  V    & 81.6          & 79.9          & 71.8          & 74.4          & 83.1                & 17.7          \\
M2R2+MSTCN     &  V,P,FT,G    & \textbf{92.6} & \textbf{89.8} & 84.6          & \textbf{93.9} & {\ul 88.9} & {\ul 70.5}    \\
M2R2+ASRF      &  V,P,FT,G    & 89.8          & 88.5          & {\ul 86.0}    & {\ul 93.7}    & 87.9        & \textbf{71.1} \\
M2R2+DiffAct   &  V,P,FT,G    & {\ul 90.5}    & {\ul 89.7}    & \textbf{89.7} & 93.3          & 83.0                  & 54.0   \\
\bottomrule
\end{tabular}
\end{table}

\begin{table}[t]
\caption{Quantitative results on the JIGSAWS dataset~\cite{gao2014jhu}. Best performing models in \textbf{bold}, second best are ~\underline{underlined}. n/r -- not reported. V -- vision, P--pose (left and right arm), G -- gripper width (left and right arm), E2E -- end-to-end.}
\label{tab:jigsaw_quant}
\centering
\scriptsize
\setlength{\tabcolsep}{2pt}
\begin{tabular}{lllllllll}
\toprule
Method   & Modalities & Type      & \multicolumn{3}{c}{F1@\{10,25,50\}}                    & Acc              & EDIT   & DR          \\
\midrule
Lea et al.~\cite{lea2016segmental}                  & V     & E2E     & n/r           & n/r           & n/r           & 74.2          & 66.5          & n/r           \\
Weerasinghe et al.~\cite{weerasinghe2024multimodal} & V,P,G & E2E     & 87.3          & 86.5          & 81.1          & 87.1          & 83.9          & n/r           \\
Atoum et al.~\cite{atoum2025multi}                  & V,P   & E2E     & n/r           & n/r           & n/r           & \textbf{90.3} & 89.0          & n/r           \\
BRP+MSTCN                                           & V     & Modular & 53.2 & 41.7 & 23.0 & 38.7 & 76.8 & 42.8 \\
BRP+ASRF                                            & V     & Modular & 53.4 & 43.2 & 24.5 & 39.3 & 73.6 & 42.3 \\
BRP+DiffAct                                         & V     & Modular & 59.6 & 49.6 & 30.2 & 44.5 & 81.9 &  43.3 \\
M2R2+MSTCN                                          & V,P,G & Modular & {\ul 93.3}    & {\ul 91.6}    & {\ul 86.3}    & 87.0          & 90.4          & 84.0          \\
M2R2+ASRF                                           & V,P,G & Modular & 93.1          & 91.2          & 86.0          & 85.6          & {\ul 90.7}          & \textbf{84.8} \\
M2R2+DiffAct                                        & V,P,G & Modular & \textbf{94.3} & \textbf{93.0} & \textbf{89.4} & {\ul 87.1}    & \textbf{91.7} & {\ul 84.5}    \\
\bottomrule
\end{tabular}
\end{table}

\begin{figure}
    \centering
    \includegraphics[width=\linewidth]{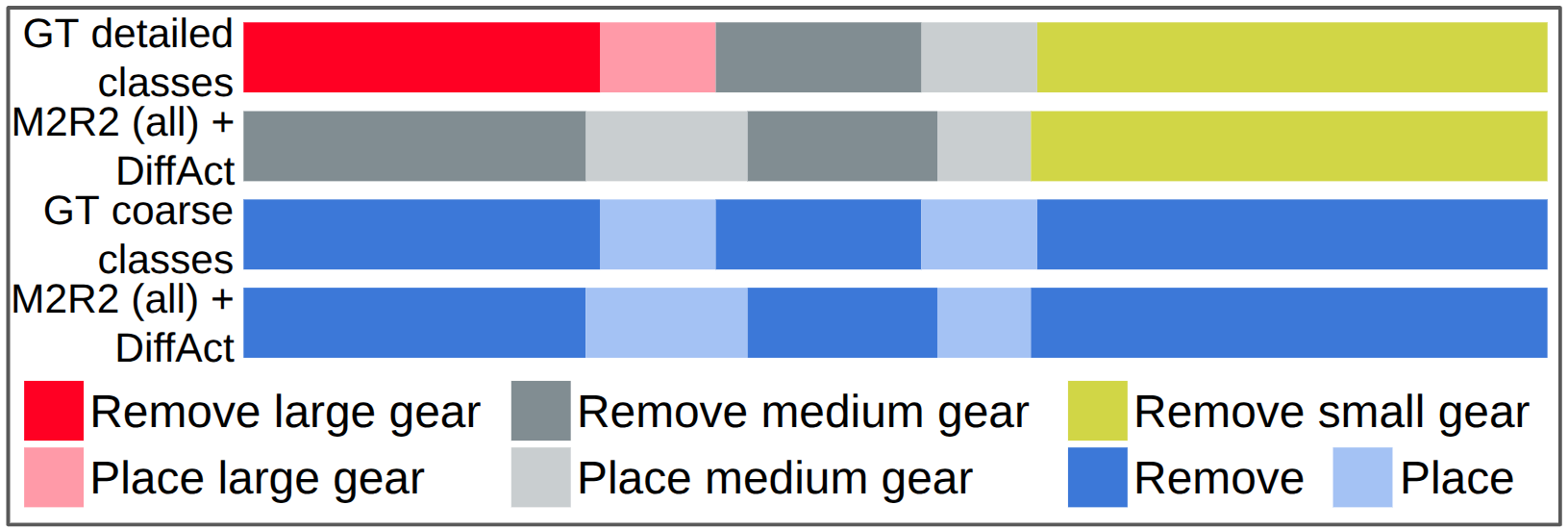}
    \caption{Coarse level prediction compared to fine-grain level prediction. Large and medium gears are confused.}
    \label{fig:coarsvdet}
\end{figure}

{
\newtext
The performance of deep learning TAS models trained with M2R2 features surpasses unsupervised approaches used in robotics. We evaluate only on the fine-grained label set, as label granularity does not affect unsupervised methods, which assign labels based on overlap with the ground truth. Due to their heuristic nature, unsupervised methods tend to over-segment demonstrations in regions with rapid sensor changes. In contrast, M2R2 avoids this issue, as the feature extractor and following TAS model learn to ignore such local variations in proprioceptive signals.
}

{
\newtext
Table~\ref{tab:ipp_quant} shows TAS performance on the (Im)PerfectPour dataset using the M2R2 extractor trained on REASSEMBLE. Our method outperforms vision-only baselines by 16.1\% in F1@50. These results indicate that when the available modalities are similar across the datasets, M2R2 features generalize well to new task domains. 

Finally, Table~\ref{tab:jigsaw_quant} reports the TAS performance of a model trained with M2R2 features compared to existing end-to-end and vision-only approaches on the JIGSAWS dataset. We outperform prior state-of-the-art methods by 1.7\% in EDIT and achieve an F1@50 of 89.4\%. Unlike previous approaches that rely on additional binary gripper signals~\cite{weerasinghe2024multimodal} or transform proprioceptive data into invariant representations~\cite{atoum2025multi}, M2R2 directly uses raw sensory readings while still achieving state-of-the-art performance. This demonstrates that by adapting the encoders to the available modalities in new embodiments and datasets, M2R2 can successfully learn meaningful multimodal representations for action segmentation.
}

\subsection{Qualitative Evaluation}
{
\newtext
Figure~\ref{fig:qual} presents qualitative results for several baseline models. TAS models with vision-only features more often confuse objects than when using multimodal features. Heuristic approaches, on the other hand, tend to over-segment demonstrations in regions with high proprioceptive variation, such as large force changes during \textit{Insert} or significant pose changes during \textit{Remove}. Even with proprioceptive data, M2R2 features occasionally confuse objects of similar size; for example, Figure~\ref{fig:coarsvdet} shows confusion between large and medium gears, as the large gear’s top diameter is similar to the medium gear’s bottom diameter. Nevertheless, coarse-level segmentation remains correct, with the model accurately identifying the \textit{Remove} and \textit{Place} sequences.
}

\begin{table}[t]
\caption{Modality ablation study on the REASSEMBLE dataset~\cite{sliwowski2025reassemble}. All results are obtained with the DiffAct TAS model~\cite{diffact}. We evaluate on fine-grain labels.}
\label{tab:abl_mod}
\centering
\begin{tabular}{lllllll}
\toprule
Method         & \multicolumn{3}{c}{F1@\{10,25,50\}}                    & Acc              & EDIT   & DR          \\
\midrule
only vision    & 27.2             & 26.1             & 21.6             & 21.4             & 28.9  & 44.3           \\
only audio     &   39.2           &  38.7           &  36.4         & 34.5            &  37.5     & 80.1      \\
only proprio      & \underline{78.0}             & \underline{77.5}             & \underline{74.5}             & \textbf{75.4}             & \textbf{69.2}  & 80.3           \\
\midrule
no gripper     & 67.9         & 67.2         & 64.2         &  65.1        & 59.2 & \underline{82.2}       \\
no pos/vel     & 69.9         & 69.1         & 65.1         &  64.6        & 61.8 & 80.5       \\
no F/T         & 72.9         & 72.4         & 69.9         &  70.4        & 65.2 & 80.6       \\
\midrule
all         & \textbf{78.1}    & \textbf{77.7}   & \textbf{74.6}    & \underline{74.9}    & \underline{68.7}  & \textbf{82.4}  \\
\bottomrule
\end{tabular}
\end{table}

\subsection{Ablation Studies}
We evaluate the impact of different sensory modalities on TAS performance by training the M2R2 feature extractor using only vision, only audio, or only proprioceptive data. We then combine both exteroceptive modalities and systematically remove specific sensors: pose and twist, gripper width, or force-torque measurements. Ablation experiments are conducted on the REASSEMBLE~\cite{sliwowski2025reassemble} dataset, which contains the widest variety of modalities.

\begin{figure}[t]
    \centering
    \begin{subfigure}[b]{0.49\textwidth}
        \centering
        \includegraphics[width=\textwidth]{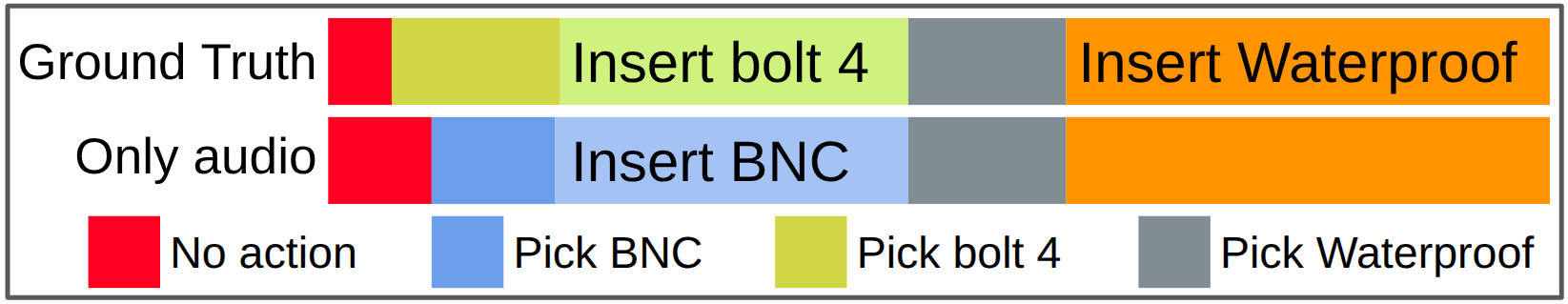}
        \caption{Some objects (like the waterproof connector) can be determined from audio alone.}
        \label{fig:sfigure1}
    \end{subfigure}
    \hfill
    \begin{subfigure}[b]{0.49\textwidth}
        \centering
        \includegraphics[width=\textwidth]{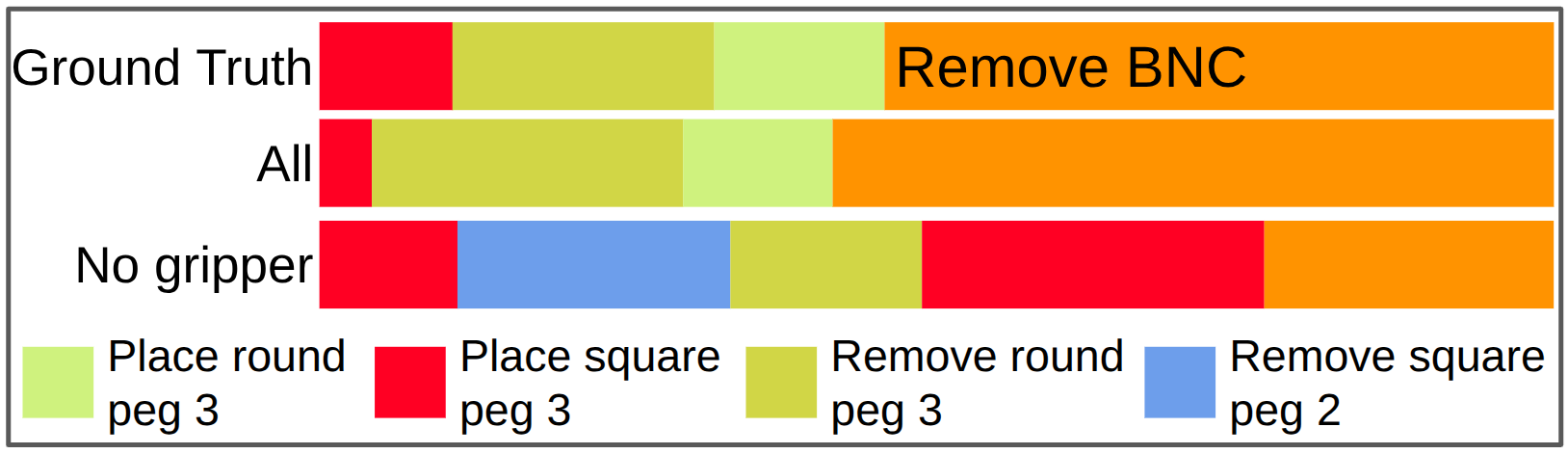}
        \caption{After removing gripper information some object categories and actions become hard to distinguish.}
        \label{fig:sfigure2}
    \end{subfigure}
    \caption{Example predictions for different modality combinations used to train the M2R2 features.}
    \label{fig:ablation_mod}
\end{figure}

Table~\ref{tab:abl_mod} shows the TAS performance with different sensory configurations. Using all modalities yields the best results, while vision alone performs worst, consistent with our earlier observation where Br-Prompt also struggles due to visibility of the objects. Audio improves boundary detection, reaching an 80.1\% detection rate, which is expected based on the REASSEMBLE dataset where transitions are often marked by distinct sounds (e.g., gripper movements, clicking of connectors, or impacts of objects). However, segmentation performance remains poor, as identifying manipulated objects from sound alone is difficult, though some objects can still be recognized (see Figure~\ref{fig:ablation_mod}).

Interestingly, TAS performance using only proprioceptive features is comparable to the model trained on all modalities. This likely reflects the specificity of the REASSEMBLE dataset, where actions and objects can often be identified from proprioceptive cues alone. Pose and twist trajectories, gripper state, and force/torque measurements help distinguish both objects and action types. However, due to the lack of diverse multimodal robotic TAS datasets, this limitation cannot be further explored; {\newtext (Im)PerfectPour is unsuitable, as it contains only one type of each object. Future work should investigate multimodal TAS in daily tasks with higher object variability, such as meal preparation, where vision is crucial for distinguishing similar objects (e.g., a bottle of orange juice vs. a bottle of water)}. Incorporating all sensory data also improves boundary placement within segmented demonstrations, as shown by the higher detection rate compared to the proprioceptive-only model.

Finally, we analyze the impact of each proprioceptive measurement on TAS performance. We observe that removing gripper information has the most significant effect on the results. This is intuitive, as objects in the REASSEMBLE dataset can often be distinguished based on their size. Figure~\ref{fig:ablation_mod} illustrates cases where the absence of gripper information makes object identification more challenging. Similarly, removing pose and twist information also significantly affects model performance, as it makes action recognition more difficult. Additionally, removing force-torque (FT) data reduces action segmentation performance for similar reasons, it becomes harder to differentiate objects.

{\newtext
Results show that some modalities impact performance more than others, partly because many objects in the REASSEMBLE dataset can be distinguished by size or mass. Prior work has shown that modalities converge at different rates~\cite{yao-mihalcea-2022-modality}, which affects the fused representation and overall performance. A common approach is to adjust the learning rates of modality encoders based on their estimated contribution~\cite{yao-mihalcea-2022-modality}.
A promising future direction is to investigate mechanisms which allow to adapt efficiently to available modalities and balance their impact on learned representations for improved action recognition performance.


}

\section{Conclusion}
In this work, we propose M2R2, a multimodal robotic representation and {\newtext training} strategy for temporal action segmentation (TAS). We leverage proprioceptive sensory information, including end-effector pose and twist, gripper width, and force-torque measurements, along with exteroceptive inputs such as vision and audio, to learn multimodal features for TAS. Unlike existing robotic multimodal TAS approaches, M2R2 features can be used with a wide range of state-of-the-art deep-learning TAS models. We compare several state-of-the-art TAS models utilizing M2R2 features against leading visual and proprioception-based TAS approaches. All baselines trained with M2R2 features outperform prior robotics and vision-based methods. {\newtext Moreover, we show strong generalization ability to new task domains and embodiments of the proposed M2R2 framework.} Additionally, we conduct an extensive ablation study to assess the impact of different modalities and sensors on TAS performance, showing that combining vision, audio, and proprioceptive data yields the highest performance.


\bibliographystyle{IEEEtran}
\bibliography{main}
\end{document}